\documentclass[letterpaper, 10 pt, journal, twoside]{IEEEtran}

\ifCLASSINFOpdf
\else
\fi
\usepackage{cite}
\usepackage{amsmath,amssymb,amsfonts}
\usepackage{graphicx}
\usepackage{textcomp}
\usepackage{booktabs} 
\usepackage{multirow}
\usepackage{array}
\usepackage[misc]{ifsym}
\usepackage{rotating}
\usepackage[small]{caption}
\usepackage{algorithm}
\usepackage[caption=false,font=normalsize,labelfont=sf,textfont=sf]{subfig}
\usepackage{url}
\usepackage{verbatim}
\usepackage{algpseudocode}   
\usepackage[breaklinks,colorlinks]{hyperref}

\usepackage{color}
\usepackage[table]{xcolor}

\newcommand{\rev}[1]{\textcolor{black}{{#1}}}

\begin{document}

\title{BEVNav: Robot Autonomous Navigation Via Spatial-Temporal Contrastive Learning in Bird's-Eye View}

\author{
 Jiahao Jiang$^{1}$, Yuxiang Yang$^{1,*}$, Yingqi Deng$^{1}$, Chenlong Ma$^{1}$, Jing Zhang$^{2}$

\thanks{$^{1}$Jiahao Jiang, Yuxiang Yang, Yingqi Deng and Chenlong Ma are with the School of Electronics and Information, Hangzhou Dianzi University, Hangzhou, China.}

\thanks{$^2$Jing Zhang is with the School of Computer Science, The University of Sydney, NSW 2006, Australia.}
}

\maketitle

\begin{abstract}

Goal-driven mobile robot navigation in map-less environments requires effective state representations for reliable decision-making. Inspired by the favorable properties of Bird's-Eye View (BEV) in point clouds for visual perception, this paper introduces a novel navigation approach named BEVNav. It employs deep reinforcement learning to learn BEV representations and enhance decision-making reliability. First, we propose a self-supervised spatial-temporal contrastive learning approach to learn BEV representations. Spatially, two randomly augmented views from a point cloud predict each other, enhancing spatial features. Temporally, we combine the current observation with consecutive frames' actions to predict future features, establishing the relationship between observation transitions and actions to capture temporal cues. Then, incorporating this spatial-temporal contrastive learning in the Soft Actor-Critic reinforcement learning framework, our BEVNav offers a superior navigation policy. Extensive experiments demonstrate BEVNav's robustness in environments with dense pedestrians, outperforming state-of-the-art methods across multiple benchmarks. \rev{The code will be made publicly available at \href{https://github.com/LanrenzzzZ/BEVNav}{BEVNav}}.

\end{abstract}

\section{Introduction}

Goal-driven mobile robot navigation in map-less environments is a fundamental and challenging task in robotics. It aims to reach a designated goal while avoiding collision in a dynamic scene. Existing methods predominantly utilize depth images to perceive the environments. For instance, \rev{Thomas et al.} \cite{thomas2021interpretable} proposes a self-attention model to extract features from depth images. \rev{de Jesus et al.} \cite{de2022depth} develops a contrastive representation learning method on depth images to guide drones in map-less navigation. \rev{Jiang et al.} \cite{jiang2023dmcl} introduces an end-to-end reinforcement learning (RL) navigation algorithm using depth images, employing depth image mask contrastive learning techniques to represent the spatial-temporal state of the scene. 

\begin{figure}[t]
\centering
\includegraphics[width=1\linewidth]{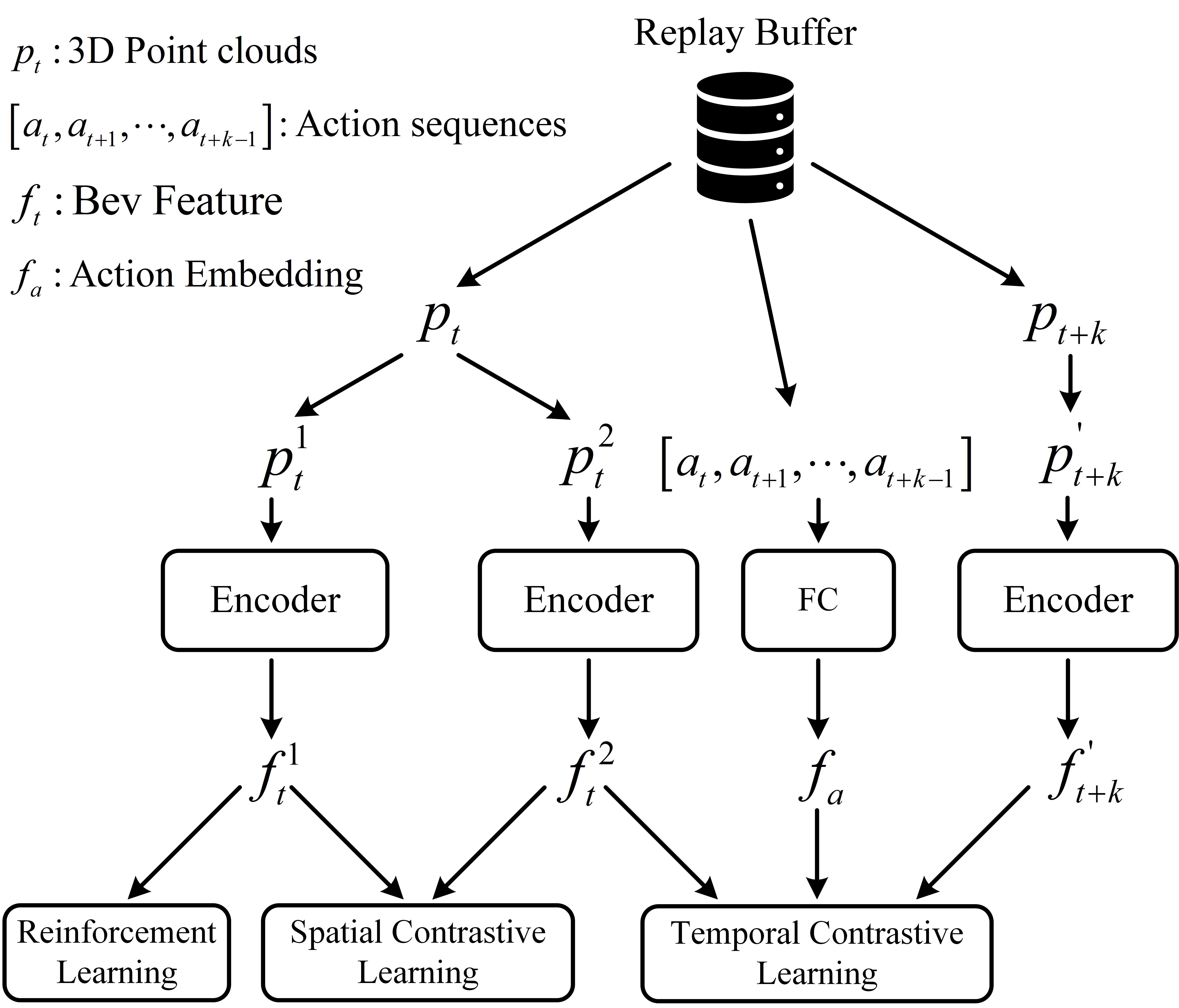}
\caption{In the BEVNav framework, we propose a Sparse-Dense BEV Network \rev{to convert 3D point clouds into BEV features efficiently}. This conversion not only creates effective scene representation but also facilitates learning effective state representations via spatial-temporal contrastive learning. As a result, this allows the alignment of these representations to the action space, thus offering a more efficient and accurate navigation policy.} 
\label{fig1}
\end{figure}

\rev{However, using depth images as 2D observations makes it challenging to directly learn the mapping relationship to 3D actions, especially for dynamic complex environments. Inspired by the success of Bird's-Eye View (BEV) representation learning for perception tasks~\cite{yang2023bevtrack,lang2019pointpillars}, we find that BEV has a significant potential to benefit 3D mobile robot navigation. BEV could better capture static and dynamic obstacles since obstacle movements largely occur in the horizontal plane in scenarios such as autonomous driving.} By compressing the 3D point clouds, BEV naturally filters out the noises in the height dimension, making it promising to plan the correct routes. To this end, we present a novel deep reinforcement learning (DRL)-based navigation approach termed BEVNav, which adopts a Sparse-Dense BEV Network to encode dense BEV features from sparse 3D point clouds via deep reinforcement learning, thus enhancing decision-making reliability.

Technically, to enhance the BEV representations, which are crucial for dynamic scene understanding and supporting making reliable decisions, we design a new self-supervised representation learning method. As shown in Fig.~\ref{fig1}, it consists of spatial contrastive learning (SCL) and temporal contrastive learning (TCL). Although RL can effectively handle decision-making problems, it does not directly address the challenge of learning effective state representation~\cite{laskin2020curl}, which is crucial in robot navigation. In contrast, self-supervised contrastive learning can significantly improve the quality of spatial representations by exploiting unlabeled observations. Inspired by SimSiam ~\cite{chen2021exploring}, we propose SCL that enables two randomly augmented views from a point cloud to predict each other with an asymmetric architecture, significantly improving the quality of the robot's visual representation. This approach provides more accurate spatial state estimation for robots in complex navigation scenarios and helps to achieve more efficient and reliable navigation performance. On the other hand, to enhance robot decision-making in complex environments, where accurately understanding and predicting the dynamic changes of obstacles in the scene is crucial for making effective navigation decisions, we propose TCL that combines the current observation with consecutive frames' actions to predict the future feature. It establishes the relationship between observation transitions and actions to capture temporal cues. Built upon these designs, BEVNav exhibits a substantial performance advantage over the current state-of-the-art (SOTA) methods in challenging scenes, including navigation in crowded pedestrian environments and generalization to unseen environments.

The main contributions of this paper are as follows:
\begin{itemize}
    \item We present BEVNav, a novel DRL-based visual navigation method that introduces the BEV representation to enhance the robot's perception of dynamic environments in the navigation domain.
    \item We introduce a Sparse-Dense BEV Network for extracting dense BEV features from sparse 3D point clouds. Additionally, we present a novel self-supervised learning approach, incorporating spatial and temporal contrastive learning. This method aids the robot in capturing both spatial cues of scene obstacles and inferring their temporal dynamics effectively.
    \item Experiments on several public benchmarks demonstrate that the proposed BEVNav outperforms previous SOTA methods, effectively improving navigation performance in challenging scenes.
\end{itemize}

\section{Related work}
\subsection{DRL in Mapless Point Goal Robot Navigation}
In the realm of robotic vision navigation, researchers are striving to enhance environmental perception and decision-making \cite{de2022depth, jiang2023dmcl, staroverov2023skill}. \rev{Further advancing the technology, Wijmans et al. \cite{wijmans2019dd} developed a distributed, decentralized, and synchronous reinforcement learning method (DD-PPO), achieving remarkable training efficiency and solving complex autonomous navigation tasks without maps. Extending this approach, Partsey et al. \cite{partsey2022mapping} optimized dataset and model sizes and used human-annotation-free data augmentation techniques to enhance navigation success in realistic PointNav challenges, even in environments lacking GPS and compass data.} Concurrently, Tsunekawa et al. \cite{lobos2020point} tackled partial observability issues with a point cloud-based method using a multi-scale feature network, although the PointNet architecture they employed did not offer clear spatial hierarchy, limiting its effectiveness in complex scenes. In contrast, this paper introduces a novel DRL-based navigation approach termed BEVNav, which converts 3D point clouds in the bird's eye view to learn effective representation for accurate visual perception and decision-making in complex environments.

\begin{figure*}[t]
\centering
\includegraphics[width=0.8\linewidth]{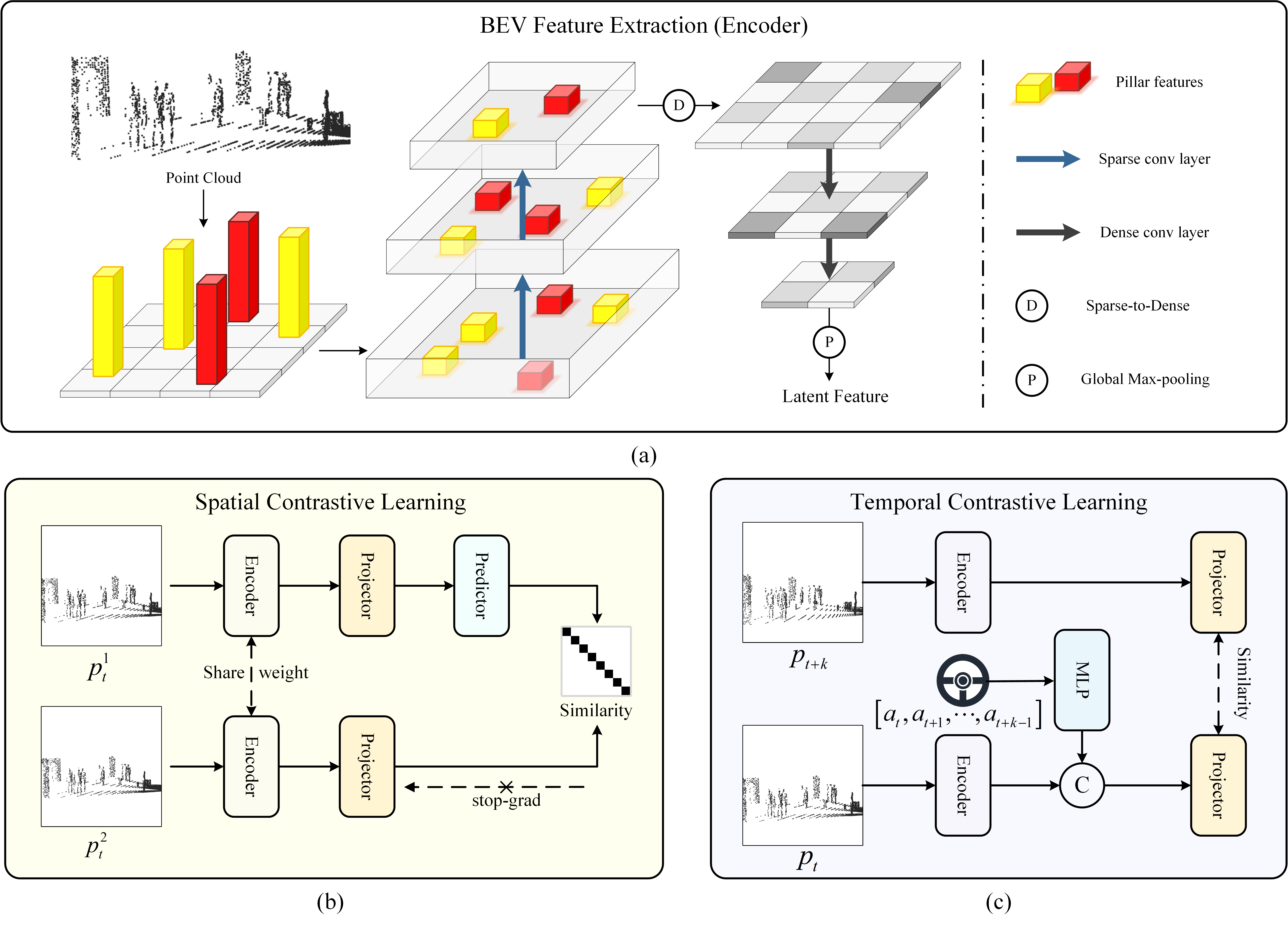}
\caption{Architecture of BEVNav for BEV feature extraction and spatial/temporal contrastive learning. We devise a new Sparse-Dense BEV network to efficiently extract BEV features from 3D point clouds, and use Global Max-pooling to obtain the latent features. Spatial contrastive learning aims to enhance the representation of spatial information by predicting data-augmented features from each other. Temporal contrastive learning aims to combine current observation with continuous frame actions to predict future features, helping to establish the relationship between observation transitions and actions.}
\label{fig2}
\end{figure*}

\subsection{3D Representations in Robot Navigation}
Robot's 3D perception modules are primarily categorized into the following three types:
(1) Multi-view projection-based models: These models extensively use images captured from different views as inputs, showing projections of the 3D environment on various image planes~\cite{guhur2023instruction,seo2023multi}. However, a significant limitation of this approach is the loss of some geometric information during the projection process.
(2) Point-based models: As in PointNet~\cite{qi2017pointnet} and PointNet++, these models directly and efficiently process 3D point clouds. In the field of robotics, many studies~\cite{fang2020graspnet,sundermeyer2021contact} have employed PointNet or PointNet++ as the encoder for visual feature extraction.
(3) Voxel-based models: These models represent another research direction to perceive 3D environments, extending the concept of 2D pixels into small cubic units in three-dimensional space~\cite{maturana2015voxnet} via voxelization. Compared to traditional point cloud or mesh models, voxelization offers a more intuitive and structured way to handle and analyze 3D point clouds. Recent works like C2FARM~\cite{james2022coarse} and PERACT~\cite{shridhar2023perceiver} have tried to use voxelized observation and action spaces for 6 degrees of freedom operations.
Different from these methods, in this paper, we introduce the BEV representation to the navigation domain and devise a Sparse-Dense BEV Network to obtain dense BEV features from sparse 3D point clouds. This representation effectively preserves spatial proximity, making the recognition and localization of objects and obstacles in complex environments more accurate and efficient.

\subsection{BEV Representation in Robot Navigation}

\rev{In the domain of robotic navigation, accurately representing environmental data, particularly through BEV, is essential for enhancing decision-making and navigational precision. Addressing the inherent challenges of this perspective, recent research has focused on developing advanced techniques to enable dynamic and precise mapping of surroundings. Li et al.\cite{li2023bi} introduced the Bi-Mapper framework to combat geometric distortions from front perspective views, integrating global and local knowledge, enhanced by an asynchronous learning policy and Across-Space Loss (ASL). Concurrently, Ross et al. \cite{ross2022bev} developed BEV-SLAM, a graph-based SLAM system using semantically-segmented BEV predictions for large-scale accurate mapping. Similarly, Liu et al. \cite{liu2023bird} proposed the BEV Scene Graph (BSG), utilizing multi-step BEV representations to surpass existing VLN methods. This paper advances these concepts by proposing a deep reinforcement learning-based BEV representation visual navigation algorithm that converts 3D point clouds into BEV and employs spatial-temporal contrast learning for efficient policy development.}

\section{Method}
This study focuses on RL for robotic autonomous navigation, aiming to learn effective policy from 3D point cloud observations through interaction with the environment. This learning process can be modeled as a Partially Observable Markov Decision Process (POMDP)~\cite{kaelbling1998planning}. In addressing POMDP challenges, RL algorithms based on the actor-critic framework have proven effective, particularly the Soft-Actor-Critic (SAC) algorithm~\cite{haarnoja2018soft}, which demonstrates superior performance. Given this, we introduce a novel DRL-based visual navigation approach termed BEVNav, which converts 3D point clouds in BEV to perceive the dynamic environments, and uses the SAC algorithm to learn navigation policy.

\subsection{Overview}
Let $M=\langle O, A, P, R,\gamma \rangle$ denote a POMDP, where $O$ represents the observation space and $A$ denotes the action space. The state transition kernel is denoted by $P:O \times A \rightarrow  \Delta O$, where $\Delta O$ signifies the distribution over the observation space. The reward function $R:O \times A \rightarrow R$ assigns immediate rewards for each observation-action pair. $\gamma $ is a discount factor, balancing the importance of immediate and future rewards. In RL, the primary objective is to find an optimal policy ${\pi ^*}:O \to \Delta \left( A \right)$ that maximizes the expected cumulative reward ${{\rm E}_\pi } = \left[ {\sum\nolimits_{t = 0}^\infty  {{\gamma ^t}{r_t}} } \right]$, focusing on long-term gain, where $\gamma \in \left[ {0,1} \right]$.

In the robot's autonomous navigation task, it is required to predict corresponding actions based on current 3D point cloud observation, aiming to reach the goal while avoiding obstacles. In this framework, let the 3D point clouds of the current frame be denoted as $p_t$, where $n_t$ represents the number of points in the 3D point clouds. At timestamp $t$, we acquire 3D point clouds through a depth camera and downsample it to 1,024 points for input. The action space consists of continuous angular and linear velocities. The action taken at time $t$ is denoted as $a_t$. Specifically, the linear velocity $v_t$ is constrained within the range of $\left( {0,1} \right)$, and the angular velocity $\omega_t$ is limited to $\left( {-1,1} \right)$. 

In RL, it is crucial to design an effective reward function to guide desired actions. Since visual navigation requires reaching the goal as quickly as possible while ensuring avoidance of collisions with any dynamic or static obstacles, we devise a multi-objective reward function in this paper:
\begin{equation}
\label{eq:reward}
r\left( {{o_t},{a_t}} \right) = \left\{ \begin{array}{l}
{r_g}{\kern 10pt}{\kern 10pt}{\kern 10pt}{\kern 10pt}{\kern 10pt}{\kern 10pt}{\kern 10pt}{\kern 10pt}{\rm{if}}{\kern 5pt}{d_t} < {\eta _D}\\
{r_c}{\kern 10pt}{\kern 10pt}{\kern 10pt}{\kern 10pt}{\kern 10pt}{\kern 10pt}{\kern 10pt}{\kern 10pt}{\rm{else {\kern 2pt}if}}{\kern 1pt}collision\\
v_t - |\omega_t | + {d_{t - 1}} - {d_t}{\kern 2pt}{\rm{otherwise}},
\end{array} \right..
\end{equation}
Eq.~\eqref{eq:reward} applies a negative reward ($r_c$) to robots that have collisions as a punishment for wrong action. On the contrary, when a robot succeeds in reaching the goal within a set time, \textit{i.e.}, when the distance to the target is lower than a predefined threshold ${\eta _D}$, it receives a positive reward ($r_g$) to encourage the correct action. In other cases, the rewards are adjusted according to the current angular and linear velocities of the robot and the variation of the distance to the goal between consecutive frames. Here, ${d_t}$ represents the distance between the robot and the goal at time $t$. In this paper, we set ${\eta _D} = 0.2$ , ${r_g} = 80$, ${r_c} = -100$.

In this study, we highlight the critical role of visual observation representation learning in navigation decision-making. As shown in Fig.~\ref{fig2}, for the BEV representation learning of 3D point clouds, our BEVNav comprises three main components: 1) \textit{BEV Feature Extraction}, 2) \textit{Spatial Contrastive Learning(SCL)}, and 3) \textit{Temporal Contrastive Learning(TCL)}. We devise a Sparse-Dense BEV Network that utilizes sparse and dense convolution to effectively embed 3D point clouds into BEV feature maps. SCL enhances the model's spatial representation capability by mutually predicting features that have undergone data-augmented. Meanwhile, TCL captures temporal clues within the scene by combining observational and action spaces to predict future features.

\subsection{BEV Feature Extraction}
To achieve a comprehensive perception of obstacles in the environment, we focus on extracting discriminative features from 3D point clouds via the Sparse-Dense BEV Network. As shown in Fig.~\ref{fig2} (a). Specifically, the point cloud is first divided into multiple vertical pillars. This not only preserves important spatial information but also converts the complex 3D point clouds into a more manageable 2D format. Given the highly sparse nature of these converted 3D point clouds, we employ sparse convolutions to extract primary spatial features from these sparse 2D pillars. During subsequent downsampling, these sparse features gradually become denser, and we utilize a series of dense convolutional networks to extract higher-level semantic features, \textit{i.e.}, BEV features $f_t^{bev} \in \mathbb{R}{^{H \times W \times C}}$. Ultimately, we extracted the latent feature $s_{t}$ through a global max-pooling layer for subsequent navigation policy learning:
\begin{equation}
{s_t} =  Pool\left( {f_{t}^{bev}} \right).
\end{equation}

\subsection{Spatial Contrastive Learning}
High-quality spatial state representation is crucial in robot navigation. Self-supervised contrastive learning can significantly improve the quality of spatial representation by utilizing unlabeled observations. In addition, it enhances the sampling efficiency in reinforcement learning~\cite{laskin2020curl}. Inspired by the success of SimSiam~\cite{chen2021exploring} in the field of self-supervised learning, we utilize an asymmetric architecture to compute the distance between predicted latent features and target features. As shown in Fig.~\ref{fig2} (b). In our approach, two randomly augmented views of the same 3D point cloud $p_t$, denoted as $p_t^1$ and $p_t^2$, serve as inputs. Each of these 3D point clouds is processed through the Sparse-Dense BEV Network to acquire their respective latent features $s_t^1$ and $s_t^2$. Among them, $s_t^1$ is sequentially processed through a projection MLP head $g$ and a prediction MLP head $h$ to generate the predicted feature ${\hat y_t} = h(g({s_t^1}))$, while $s_t^2$ undergoes processed only through the projection MLP head $g$, culminating in the formation of the target feature ${\bar y_t} = g({s_t^2})$. SCL aims to make the predictive feature as close as possible to the target feature. To achieve this, we adopt a loss function based on cosine similarity between ${\bar y_t}$ and ${\hat y_t}$, which can be formulated below:
\begin{equation}
{L_{sc}} = 1 - \frac{1}{N}\sum\limits_{i = 0}^N {\frac{{\hat y_{t+i}}}{{{{\left\| {\hat y_{t+i}} \right\|}_2}}}}  \cdot \frac{{\bar y_{t+i}}}{{{{\left\| {\bar y_{t+i}} \right\|}_2}}}.
\end{equation}

\subsection{Temporal Contrastive Learning}
To enhance the robot's dynamic reasoning ability in complex environments, accurately predicting changes in the scene is crucial for making effective navigation decisions. As shown in Fig.~\ref{fig2} (c), the core of TCL lies in combining current observations with a series of consecutive action frames, to predict future features and establish the relationship between observation transitions and actions to capture temporal cues. In this way, it achieves an accurate alignment between the state representation and the action space. To this end, we randomly sample a batch of transitions $\left\{ {p_{_t}^i,[a_{_t}^i, a_{_{t + 1}}^i,..., a_{_{t + k}}^i],p_{_{t + k}}^i} \right\}$ from the state observations and action sequences. \rev{In this process, we input the observations at times $t$ and $t+k$ into the Sparse-Dense BEV Network to obtain the corresponding BEV features $f_t^{bev}$ and $f_{t+k}^{bev}$ and latent feature $s_t$, ${s_{t + k}}$.}

\rev{Subsequently, the action sequence is encoded using a lightweight MLP and concatenated with the latent feature $s_t$. This combined feature is then passed through another lightweight MLP to predict the features $\hat x_{t + k}$. Simultaneously, ${s_{t + k}}$ is processed through an MLP to obtain the feature $x_{t + k}$. \(\hat{x}_{t+k}\) and $x_{t + k}$ are then learned through spatial-temporal contrastive learning.} To optimize this process, we employ a contrastive loss function ${L_{tc}}$ for training. 
\begin{equation}
{c} = MLP\left( {\left[ {a_{_t},a_{t + 1},...a_{t + k}} \right]} \right),
\end{equation}
\begin{equation}
{\hat x_{t + k}} = MLP\left( {[c,{s_t}]} \right),
\end{equation}
\begin{equation}
{x_{t + k}} =  MLP\left(  {s_{t + k}} \right),
\end{equation}
\begin{equation}
{L_{tc}} =  - \frac{1}{N}\sum\limits_{i = 1}^N {\log \frac{{\hat x_{t + k}^i \cdot x_{t + k}^i}}{{\sum\nolimits_j^N {\hat x_{t + k}^i \cdot x_{t + k}^j} }}}.
\end{equation}

\begin{figure}[t]
\centering
\includegraphics[width=0.9\linewidth]{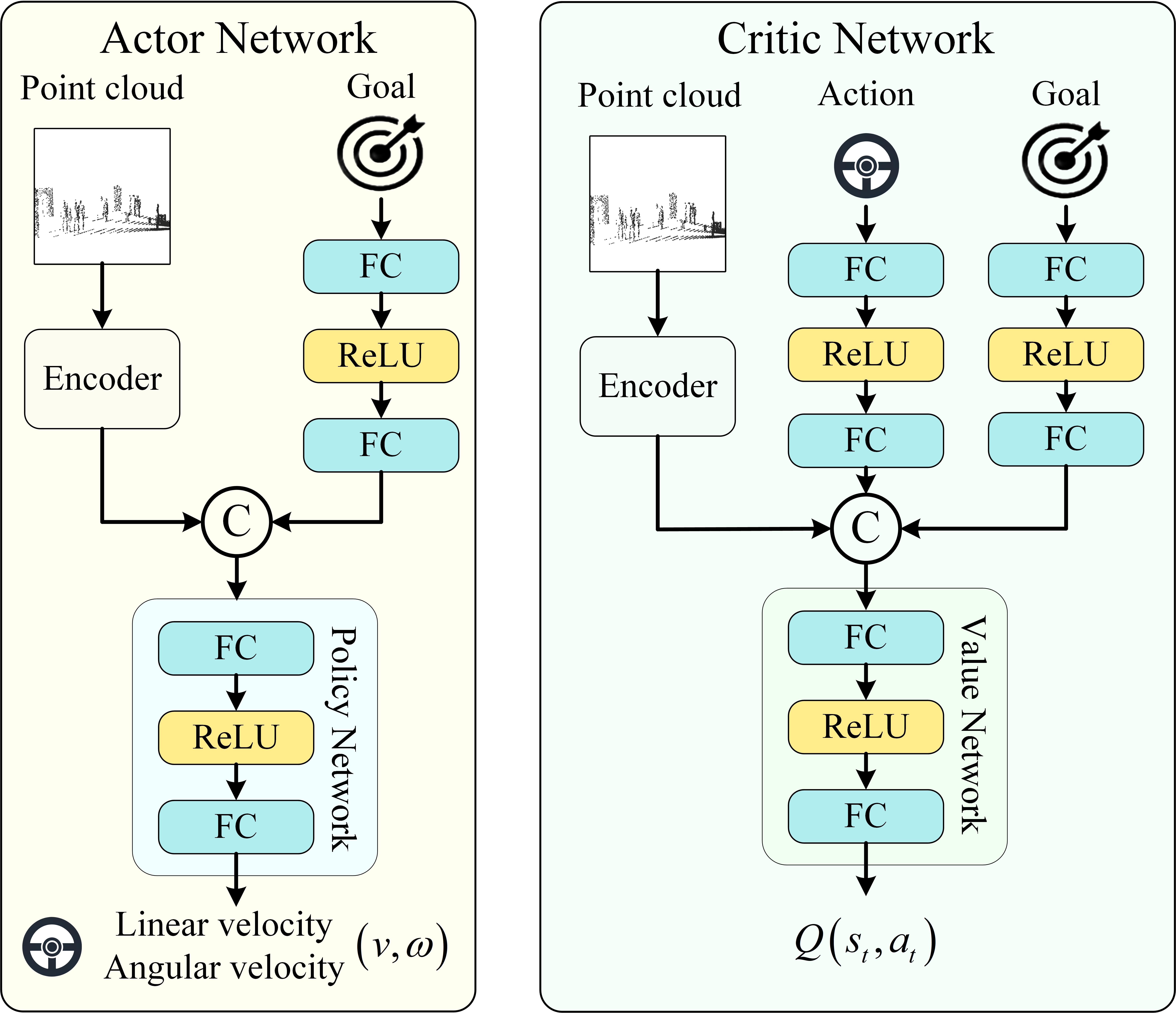}
\caption{SAC-based Navigation policy learning framework. It compromises two key components: 1) BEV feature extraction, and 2) BEV-based action decision-making and action evaluation.} 
\label{fig3}
\end{figure}

\subsection{SAC-based Reinforcement Learning}
Although many algorithms exist in the field of DRL for learning navigation policy~\cite{hu2021sim, han2022reinforcement, josef2020deep, yang2023st}, we chose the SAC algorithm to train our DRL network, as shown in Fig.~\ref{fig3}. Specifically, the Actor Network consists of a Spare-Dense BEV Network and a policy network, which combines BEV features and the goal of navigation action decision-making. The Critic Network is composed of the Spare-Dense BEV Network and a Q value network, which combine BEV features, goals, and actions for action evaluation to assess the quality of actions. A significant advantage of the SAC algorithm is its excellent sample efficiency and stable learning performance, making it particularly effective in complex environments. Additionally, SAC enhances the policy's exploration capabilities by maximizing both the expected reward and the entropy of the action, often leading to outcomes comparable to or better than other SOTA algorithms(\textit{e.g.}, A2C ~\cite{mnih2016asynchronous} and TRPO ~\cite{schulman2015trust}).

\subsection{Implementation Details}

\rev{In our study, the 3D point clouds are sourced from the RGBD camera. The ranges of the point cloud input in the x, y, and z directions are (-9.6, 9.6), (-1.6, 0.448), and (0, 10) meters, respectively, with pillar sizes of $[0.15, 0.016, 0.5]$.} 
\rev{It’s important to note that the height of each pillar matches the range of the point cloud in the z-direction, which results in the 3D point cloud being compressed along the z-axis and transformed into a BEV feature.} For feature extraction, we utilize a Sparse-Dense BEV network backbone composed of four sparse convolution blocks and three dense convolution blocks, where the sparse blocks consist of $\left\{ {{\rm{16}},{\rm{32}},{\rm{64}},{\rm{128}}} \right\}$ channels and the dense blocks have 128 channels each. 

\begin{figure}[!t]
\centering
\includegraphics[width=0.9\linewidth]{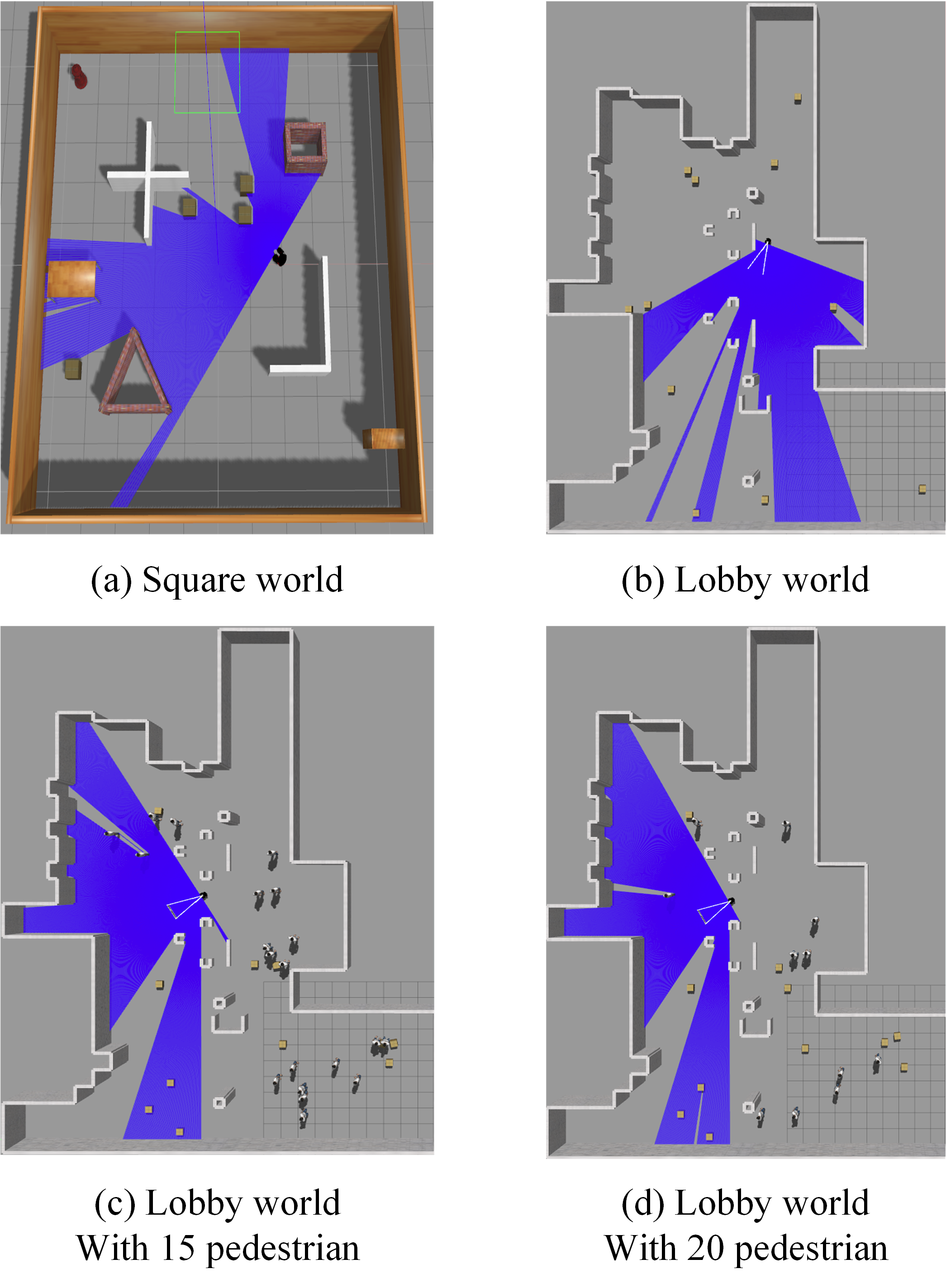}
\caption{Gazebo simulation environments: the Square-World and the Lobby-World. The Square-World offers a vast open space, used for training and testing. In contrast, the Lobby-World represents a more complex and dynamic environment, only used for testing and optimizing robotic navigation policy in crowded settings.} 
\label{fig4}
\end{figure}

\section{Experiment}
In this section, we provide a detailed description of our experimental setup and conduct a series of experiments aimed at demonstrating the performance superiority and practical efficacy of our method compared to other SOTA approaches. 

\subsection{Simulation Setup}
In our study, we utilize the Gazebo simulator~\cite{koenig2004design} and the PEDSIM library\footnote{http://pedsim.silmaril.org}\footnote{https://github.com/srl-freiburg/pedsim} to conduct simulation experiments. The robot is equipped with an Intel Realsense D435i RGB-D camera sensor, which captures real-time information about dynamic and static obstacles in the scene using 3D point clouds. The depth of the D435i camera is set to $\left[ {0.3,10} \right]$m, and its FOV is about ${85^ \circ }$. \rev{The speed of pedestrians is set as 1 meter per second. We employed two different experimental scenarios ~\cite{cimurs2021goal,xie2023drl}, as shown in Fig.~\ref{fig4}.} During the training process, each episode randomly generates the location of boxed obstacles as well as the robot's initial location and goal within the map-free space. 

\subsection{Training Details}
We implement the algorithm using the PyTorch framework. The model parameters are optimized using the Adam optimizer~\cite{kingma2014adam} with the learning rate decaying in steps from $10^{-3}$, and the batch size is 64. Furthermore, we employ a data augmentation technique involving random shifts \cite{ling2023efficacy} of magnitude 0.01 during training.

\subsection{Main Results and Analysis}
For comparison, we employ three widely used metrics in navigation literature: 1) Success Rate (SR): defined as $\frac{1}{N}\sum\limits_{i = 1}^N {{S_i}}$ where $N=100$ is the number of episodes and $S_i$ is a binary indicator of success in episodes; 2) Navigation velocity; 3) Success Path Length (SPL): defined as $\frac{1}{N}\sum\limits_{i = 1}^N {{S_i}} \frac{{{L_i}}}{{\max \left( {{P_i},{L_i}} \right)}}$ which measures the path quality and the navigation efficiency when the robot successfully achieves the goal in the episode $i$, where $P_i$ is the path length, and $L_i$ is the optimal trajectory length. \rev{We also give the comparison of “reward”, which can offer valuable insights into model behavior during navigation.}

\begin{table}[!h]
\centering
\caption{\rev{Results in the environments without pedestrians.}}
\footnotesize
\begin{tabular}{cccccc|c}
\toprule
World & Method & Modality & SR  &  Velocity  & SPL  & Reward  \\ \midrule
\multirow{3}{*}{Square} & DMCL\cite{jiang2023dmcl}  & Image  & 0.89   & 0.75   & 0.83    & 66.51     \\
                            & SAC-P\cite{lobos2020point}     & Pcd   & 0.85    & 0.73     & 0.80   & 60.32     \\     
                            & SAC-B    & Pcd    & 0.88   & 0.74  & 0.81 & 65.82 \\
\multirow{1}{*}{World}      & CURL-B   & Pcd    & 0.90  & 0.74   & 0.83 & 68.48 
                            \\
                            & BEVNav   & Pcd   & \textbf{0.95}  & \textbf{0.76}  & \textbf{0.85}  & \textbf{77.54} \\ \midrule
\multirow{1}{*}{Lobby}      & DMCL\cite{jiang2023dmcl}  & Image & 0.87  & 0.75 
                            & 0.83   & 62.58   \\
\multirow{2}{*}{World}      & SAC-P\cite{lobos2020point}  & Pcd  & 0.83  & 0.73                             & 0.80  & 55.64 \\
                            & BEVNav   & Pcd   & \textbf{0.94}  & \textbf{0.76} & \textbf{0.85}  & \textbf{75.34} \\ \bottomrule
\end{tabular}
\label{tab1}
\end{table}

We first conduct comparisons in the environments without pedestrians, as shown in Table~\ref{tab1}. \rev{SAC-P and SAC-B represent two baselines using soft actor-critic~\cite{haarnoja2018soft} with different feature extraction techniques, \textit{i.e.}, the traditional PointNet~\cite{lobos2020point} and our Sparse-Dense BEV Network, respectively. CURL-B is built upon the SAC-B by adopting spatial contrastive learning~\cite{laskin2020curl}. DMCL~\cite{jiang2023dmcl} is developed based on spatial-temporal masked contrastive learning with depth images as state observation. Limited by the use of depth images, the performance of DMCL is unsatisfactory. The Sparse-Dense BEV Network method SAC-B outperforms the PointNet method SAC-P in terms of navigation success rate and efficiency. This validates the effectiveness of our Sparse-Dense BEV network, affirming that transforming 3D point clouds into BEV representation helps perceive the environment better and is more suitable for the robot navigation task. CURL-B adopts spatial contrastive learning as an auxiliary task, obtaining better performance. Finally, the results of our BEVNav outperform previous methods by combining temporal and spatial contrastive learning auxiliary tasks. Moreover, we test the models in the Lobby-World environment, which is unseen during training. The results in the bottom part of Table~\ref{tab1} indicate that our BEVNav maintains excellent success rates and efficiency, confirming its superior generalization performance.}



\begin{table}[htbp]
\centering
\caption{\rev{Results on different crowded-Pedestrian settings.}}
\footnotesize
\begin{tabular}{cccccc|c}
\toprule
World & Method & Modality & SR  & Velocity  & SPL & Reward \\ \midrule
\multirow{1}{*}{Lobby}                 & SAC-P\cite{lobos2020point}    & Pcd      & 0.76    & 0.73      & 0.80    & 41.11    \\
\multirow{1}{*}{World}                  & DMCL\cite{jiang2023dmcl}     & Image    & 0.82    & 0.75      & 0.82   & 52.64   \\
\multirow{1}{*}{5 ped}     & BEVNav  & Pcd    & \textbf{0.87}    & \textbf{0.76}     & \textbf{0.84}   & \textbf{62.87}  \\ \midrule
\multirow{1}{*}{Lobby}     & SAC-P\cite{lobos2020point}   & Pcd    & 0.70  & 0.71    & 0.78     & 28.21 \\
\multirow{1}{*}{World}      & DMCL\cite{jiang2023dmcl}   & Image    & 0.76   & 0.74          & 0.80    & 42.35 \\
\multirow{1}{*}{10 ped}      & BEVNav   & Pcd   & \textbf{0.83}   & \textbf{0.74}   & \textbf{0.83}  & \textbf{54.67} \\ \midrule
\multirow{1}{*}{Lobby}     & SAC-P\cite{lobos2020point}     & Pcd   & 0.63    & 0.70      & 0.76    & 15.68 \\
\multirow{1}{*}{World}                  & DMCL\cite{jiang2023dmcl}        & Image    & 0.70    & 0.72    & 0.78    & 27.35      \\
\multirow{1}{*}{15 ped}      & BEVNav     & Pcd      & \textbf{0.78}   & \textbf{0.73}     & \textbf{0.82}   & \textbf{45.67}  \\ \midrule       
\multirow{1}{*}{Lobby}        & SAC-P\cite{lobos2020point}     & Pcd      & 0.55    & 0.68     & 0.72    & 2.63   \\
\multirow{1}{*}{World}                  & DMCL\cite{jiang2023dmcl}        & Image    & 0.68    & 0.70      & 0.75     & 24.35   \\
\multirow{1}{*}{20 ped}    & BEVNav   & Pcd    & \textbf{0.75}     & \textbf{0.72}      & \textbf{0.80}   & \textbf{40.68}  \\ \bottomrule  
\end{tabular}
\label{tab2}
\end{table}

\rev{Then, we perform comparisons in the Lobby-World with 5-20 pedestrians (sampling interval 5). All the models are only trained in the Square-World environment. As shown in Table ~\ref{tab2}, SAC-P exhibits subpar navigation performance in environments with pedestrians. Benefit from spatial-temporal state representation, DMCL achieves better performance than SAC-P. But, restricted by using depth image as state observation, the performance of DMCL is inferior to our BEVNav. The results of our BEVNav demonstrate the superiority of the proposed BEV representation and spatial-temporal contrastive learning.} The strong generalization ability of our policy in unseen complex environments is also highlighted. It is also worth noting that the decrease in success rate is partly due to the limited field of view of the robot, e.g., the inability to detect pedestrians behind it, making it challenging to avoid them effectively.




\subsection{Ablation Study}

\begin{table}[htbp]
\centering
\caption{Ablation study of different prediction windows.}
\footnotesize
\begin{tabular}{ccccc|c}
\toprule
World & Prediction Window  & SR & Velocity & SPL & Reward \\ \midrule
\multirow{2}{*}{Square}      & 0     & 0.90    & 0.74   & 0.83   & 68.12  \\
                             & 1     & 0.92    & 0.75   & 0.84  & 71.24  \\
\multirow{1}{*}{World}      & 2    & 0.94    & 0.75    & 0.84   & 75.34 \\
                            & 3     & 0.95   & 0.76    & 0.85  & 77.54   \\ \bottomrule 
\end{tabular}
\label{tab3}
\end{table}

In addition to the investigation of the design choice presented in Table~\ref{tab1}, we also study the impact of the prediction window $K$ in the temporal conservative learning. This hyper-parameter plays a critical role in enhancing navigation performance. In our study, we investigated three different settings of $K$ and observed improvements with the increase of $K$ from 1 to 3. As shown in Table~\ref{tab3}, when $K=1$, there is a significant performance improvement compared to the result without using the temporal contrastive learning auxiliary task. Given the complexity and variability of scenarios, larger $K$ values can provide more accurate predictions. However, the performance gains at $K=2$ and $K=3$ are not as pronounced as at $K=1$. Based on these findings, we finally select $K=3$ as the default setting. It is important to note that the optimal value of $K$ may vary for other continuous control tasks.

\subsection{Discussion}
While the proposed BEVNav performs competitively in challenging scenarios, it may be subject to collisions when the number of pedestrians increases. More research efforts could be made in several directions, \textit{e.g.}, incorporating temporal and multi-modal information. In our experiments, we use only the point cloud of the current frame as state observation, without relying on temporal information present in previous frames. It is promising to effectively predict pedestrian motion trajectories using both the current frame and previous frames. For example, temporal information ~\cite{li2022bevformer} can be readily introduced in our BEVNav, by connecting additional temporal BEV features from previous frames. \rev{Moreover, another limitation of our method is how to define navigation goal and calculate the distance in real-world mapless scenarios. In future research, we can utilize RTK-GPS for outdoor navigation and various indoor localization systems employing fiducial markers, to bridge the gap between simulated environments and real-world applications.} Furthermore, point clouds are typically sparse and incomplete lacking scene semantic information, while image captures scene details and improves rich semantic information for scene representation. BEV, as a unified representation for multi-modal fusion ~\cite{liu2023bevfusion}, adeptly preserves semantic information from images and geometric information from point clouds. We envision that BEVNav can inspire further research into multi-frame, multi-modal fusion methodologies for robot visual navigation. 

\section{Conclusion}
This work presents a novel deep reinforcement learning (DRL)-based visual navigation approach, \textit{i.e.}, BEVNav. It introduces the Bird's-Eye View (BEV) representation to enhance the robot's perception of dynamic environments in the navigation domain. Specifically, we design a Sparse-Dense BEV Network as an encoder, which extracts the BEV features of the 3D point cloud to perceive the scene obstacles effectively. In addition, we design the Spatial-Temporal Contrastive learning in the RL framework, which helps to learn better spatial features and capture temporal cues via establishing the relationship between observation transitions and actions. Extensive experiments have demonstrated that BEVNav can achieve high-quality navigation in a variety of unseen, complex, and crowded pedestrian environments.

\bibliographystyle{IEEEtran}
\bibliography{ref-iral}

\begin{thebibliography}{10}
\providecommand{\url}[1]{#1}
\csname url@samestyle\endcsname
\providecommand{\newblock}{\relax}
\providecommand{\bibinfo}[2]{#2}
\providecommand{\BIBentrySTDinterwordspacing}{\spaceskip=0pt\relax}
\providecommand{\BIBentryALTinterwordstretchfactor}{4}
\providecommand{\BIBentryALTinterwordspacing}{\spaceskip=\fontdimen2\font plus
\BIBentryALTinterwordstretchfactor\fontdimen3\font minus \fontdimen4\font\relax}
\providecommand{\BIBforeignlanguage}[2]{{%
\expandafter\ifx\csname l@#1\endcsname\relax
\typeout{** WARNING: IEEEtran.bst: No hyphenation pattern has been}%
\typeout{** loaded for the language `#1'. Using the pattern for}%
\typeout{** the default language instead.}%
\else
\language=\csname l@#1\endcsname
\fi
#2}}
\providecommand{\BIBdecl}{\relax}
\BIBdecl

\bibitem{thomas2021interpretable}
D.-G. Thomas, D.~Olshanskyi, K.~Krueger, T.~Wongpiromsarn, and A.~Jannesari, ``Interpretable uav collision avoidance using deep reinforcement learning,'' \emph{arXiv preprint arXiv:2105.12254}, 2021.

\bibitem{de2022depth}
J.~C. de~Jesus, V.~A. Kich, A.~H. Kolling, R.~B. Grando, R.~S. Guerra, and P.~L. Drews, ``Depth-cuprl: Depth-imaged contrastive unsupervised prioritized representations in reinforcement learning for mapless navigation of unmanned aerial vehicles,'' in \emph{2022 IEEE/RSJ International Conference on Intelligent Robots and Systems (IROS)}, 2022, pp. 10\,579--10\,586.

\bibitem{jiang2023dmcl}
J.~Jiang, P.~Li, X.~Lv, and Y.~Yang, ``Dmcl: Robot autonomous navigation via depth image masked contrastive learning,'' in \emph{2023 IEEE/RSJ International Conference on Intelligent Robots and Systems (IROS)}, 2023, pp. 5172--5178.

\bibitem{yang2023bevtrack}
Y.~Yang, Y.~Deng, J.~Zhang, J.~Nie, and Z.-J. Zha, ``Bevtrack: A simple and strong baseline for 3d single object tracking in bird's-eye view,'' \emph{arXiv e-prints}, pp. arXiv--2309, 2023.

\bibitem{lang2019pointpillars}
A.~H. Lang, S.~Vora, H.~Caesar, L.~Zhou, J.~Yang, and O.~Beijbom, ``Pointpillars: Fast encoders for object detection from point clouds,'' in \emph{Proceedings of the IEEE/CVF Conference on Computer Vision and Pattern Recognition}, 2019, pp. 12\,697--12\,705.

\bibitem{laskin2020curl}
M.~Laskin, A.~Srinivas, and P.~Abbeel, ``Curl: Contrastive unsupervised representations for reinforcement learning,'' in \emph{International Conference on Machine Learning}.\hskip 1em plus 0.5em minus 0.4em\relax PMLR, 2020, pp. 5639--5650.

\bibitem{chen2021exploring}
X.~Chen and K.~He, ``Exploring simple siamese representation learning,'' in \emph{Proceedings of the IEEE/CVF Conference on Computer Vision and Pattern Recognition}, 2021, pp. 15\,750--15\,758.

\bibitem{staroverov2023skill}
A.~Staroverov, K.~Muravyev, K.~Yakovlev, and A.~I. Panov, ``Skill fusion in hybrid robotic framework for visual object goal navigation,'' \emph{Robotics}, vol.~12, no.~4, p. 104, 2023.

\bibitem{wijmans2019dd}
E.~Wijmans, A.~Kadian, A.~Morcos, S.~Lee, I.~Essa, D.~Parikh, M.~Savva, and D.~Batra, ``Dd-ppo: Learning near-perfect pointgoal navigators from 2.5 billion frames,'' \emph{arXiv preprint arXiv:1911.00357}, 2019.

\bibitem{partsey2022mapping}
R.~Partsey, E.~Wijmans, N.~Yokoyama, O.~Dobosevych, D.~Batra, and O.~Maksymets, ``Is mapping necessary for realistic pointgoal navigation?'' in \emph{Proceedings of the IEEE/CVF Conference on Computer Vision and Pattern Recognition}, 2022, pp. 17\,232--17\,241.

\bibitem{lobos2020point}
K.~Lobos-Tsunekawa and T.~Harada, ``Point cloud based reinforcement learning for sim-to-real and partial observability in visual navigation,'' in \emph{2020 IEEE/RSJ International Conference on Intelligent Robots and Systems (IROS)}, 2020, pp. 5871--5878.

\bibitem{guhur2023instruction}
P.-L. Guhur, S.~Chen, R.~G. Pinel, M.~Tapaswi, I.~Laptev, and C.~Schmid, ``Instruction-driven history-aware policies for robotic manipulations,'' in \emph{Conference on Robot Learning}.\hskip 1em plus 0.5em minus 0.4em\relax PMLR, 2023, pp. 175--187.

\bibitem{seo2023multi}
Y.~Seo, J.~Kim, S.~James, K.~Lee, J.~Shin, and P.~Abbeel, ``Multi-view masked world models for visual robotic manipulation,'' \emph{arXiv preprint arXiv:2302.02408}, 2023.

\bibitem{qi2017pointnet}
C.~R. Qi, H.~Su, K.~Mo, and L.~J. Guibas, ``Pointnet: Deep learning on point sets for 3d classification and segmentation,'' in \emph{Proceedings of the IEEE Conference on Computer Vision and Pattern Recognition}, 2017, pp. 652--660.

\bibitem{fang2020graspnet}
H.-S. Fang, C.~Wang, M.~Gou, and C.~Lu, ``Graspnet-1billion: A large-scale benchmark for general object grasping,'' in \emph{Proceedings of the IEEE/CVF Conference on Computer Vision and Pattern Recognition}, 2020, pp. 11\,444--11\,453.

\bibitem{sundermeyer2021contact}
M.~Sundermeyer, A.~Mousavian, R.~Triebel, and D.~Fox, ``Contact-graspnet: Efficient 6-dof grasp generation in cluttered scenes,'' in \emph{2021 IEEE International Conference on Robotics and Automation (ICRA)}, 2021, pp. 13\,438--13\,444.

\bibitem{maturana2015voxnet}
D.~Maturana and S.~Scherer, ``Voxnet: A 3d convolutional neural network for real-time object recognition,'' in \emph{2015 IEEE/RSJ international conference on intelligent robots and systems (IROS)}, 2015, pp. 922--928.

\bibitem{james2022coarse}
S.~James, K.~Wada, T.~Laidlow, and A.~J. Davison, ``Coarse-to-fine q-attention: Efficient learning for visual robotic manipulation via discretisation,'' in \emph{Proceedings of the IEEE/CVF Conference on Computer Vision and Pattern Recognition}, 2022, pp. 13\,739--13\,748.

\bibitem{shridhar2023perceiver}
M.~Shridhar, L.~Manuelli, and D.~Fox, ``Perceiver-actor: A multi-task transformer for robotic manipulation,'' in \emph{Conference on Robot Learning}.\hskip 1em plus 0.5em minus 0.4em\relax PMLR, 2023, pp. 785--799.

\bibitem{li2023bi}
S.~Li, K.~Yang, H.~Shi, J.~Zhang, J.~Lin, Z.~Teng, and Z.~Li, ``Bi-mapper: Holistic bev semantic mapping for autonomous driving,'' \emph{IEEE Robotics and Automation Letters}, 2023.

\bibitem{ross2022bev}
J.~Ross, O.~Mendez, A.~Saha, M.~Johnson, and R.~Bowden, ``Bev-slam: Building a globally-consistent world map using monocular vision,'' in \emph{2022 IEEE/RSJ International Conference on Intelligent Robots and Systems (IROS)}.\hskip 1em plus 0.5em minus 0.4em\relax IEEE, 2022, pp. 3830--3836.

\bibitem{liu2023bird}
R.~Liu, X.~Wang, W.~Wang, and Y.~Yang, ``Bird's-eye-view scene graph for vision-language navigation,'' in \emph{Proceedings of the IEEE/CVF International Conference on Computer Vision}, 2023, pp. 10\,968--10\,980.

\bibitem{kaelbling1998planning}
L.~P. Kaelbling, M.~L. Littman, and A.~R. Cassandra, ``Planning and acting in partially observable stochastic domains,'' \emph{Artificial intelligence}, vol. 101, no. 1-2, pp. 99--134, 1998.

\bibitem{haarnoja2018soft}
T.~Haarnoja, A.~Zhou, P.~Abbeel, and S.~Levine, ``Soft actor-critic: Off-policy maximum entropy deep reinforcement learning with a stochastic actor,'' in \emph{International conference on Machine Learning}.\hskip 1em plus 0.5em minus 0.4em\relax PMLR, 2018, pp. 1861--1870.

\bibitem{hu2021sim}
H.~Hu, K.~Zhang, A.~H. Tan, M.~Ruan, C.~Agia, and G.~Nejat, ``A sim-to-real pipeline for deep reinforcement learning for autonomous robot navigation in cluttered rough terrain,'' \emph{IEEE Robotics and Automation Letters}, vol.~6, no.~4, pp. 6569--6576, 2021.

\bibitem{han2022reinforcement}
R.~Han, S.~Chen, S.~Wang, Z.~Zhang, R.~Gao, Q.~Hao, and J.~Pan, ``Reinforcement learned distributed multi-robot navigation with reciprocal velocity obstacle shaped rewards,'' \emph{IEEE Robotics and Automation Letters}, vol.~7, no.~3, pp. 5896--5903, 2022.

\bibitem{josef2020deep}
S.~Josef and A.~Degani, ``Deep reinforcement learning for safe local planning of a ground vehicle in unknown rough terrain,'' \emph{IEEE Robotics and Automation Letters}, vol.~5, no.~4, pp. 6748--6755, 2020.

\bibitem{yang2023st}
Y.~Yang, J.~Jiang, J.~Zhang, J.~Huang, and M.~Gao, ``S{T}2: Spatial-temporal state transformer for crowd-aware autonomous navigation,'' \emph{IEEE Robotics and Automation Letters}, 2023.

\bibitem{mnih2016asynchronous}
V.~Mnih, A.~P. Badia, M.~Mirza, A.~Graves, T.~Lillicrap, T.~Harley, D.~Silver, and K.~Kavukcuoglu, ``Asynchronous methods for deep reinforcement learning,'' in \emph{International Conference on Machine Learning}.\hskip 1em plus 0.5em minus 0.4em\relax PMLR, 2016, pp. 1928--1937.

\bibitem{schulman2015trust}
J.~Schulman, S.~Levine, P.~Abbeel, M.~Jordan, and P.~Moritz, ``Trust region policy optimization,'' in \emph{International Conference on Machine Learning}.\hskip 1em plus 0.5em minus 0.4em\relax PMLR, 2015, pp. 1889--1897.

\bibitem{koenig2004design}
N.~Koenig and A.~Howard, ``Design and use paradigms for gazebo, an open-source multi-robot simulator,'' in \emph{2004 IEEE/RSJ International Cconference on Intelligent Robots and Systems (IROS)}, vol.~3, 2004, pp. 2149--2154.

\bibitem{cimurs2021goal}
R.~Cimurs, I.~H. Suh, and J.~H. Lee, ``Goal-driven autonomous exploration through deep reinforcement learning,'' \emph{IEEE Robotics and Automation Letters}, vol.~7, no.~2, pp. 730--737, 2021.

\bibitem{xie2023drl}
Z.~Xie and P.~Dames, ``Drl-vo: Learning to navigate through crowded dynamic scenes using velocity obstacles,'' \emph{IEEE Transactions on Robotics}, 2023.

\bibitem{kingma2014adam}
D.~P. Kingma and J.~Ba, ``Adam: A method for stochastic optimization,'' \emph{arXiv preprint arXiv:1412.6980}, 2014.

\bibitem{ling2023efficacy}
Z.~Ling, Y.~Yao, X.~Li, and H.~Su, ``On the efficacy of 3d point cloud reinforcement learning,'' \emph{arXiv preprint arXiv:2306.06799}, 2023.

\bibitem{li2022bevformer}
Z.~Li, W.~Wang, H.~Li, E.~Xie, C.~Sima, T.~Lu, Y.~Qiao, and J.~Dai, ``Bevformer: Learning bird’s-eye-view representation from multi-camera images via spatiotemporal transformers,'' in \emph{European conference on computer vision}.\hskip 1em plus 0.5em minus 0.4em\relax Springer, 2022, pp. 1--18.

\bibitem{liu2023bevfusion}
Z.~Liu, H.~Tang, A.~Amini, X.~Yang, H.~Mao, D.~L. Rus, and S.~Han, ``Bevfusion: Multi-task multi-sensor fusion with unified bird's-eye view representation,'' in \emph{2023 IEEE International Conference on Robotics and Automation (ICRA)}, 2023, pp. 2774--2781.

\end{thebibliography}

\end{document}